\documentclass[10pt,twocolumn,letterpaper]{article}

\usepackage{cvpr}              

\definecolor{cvprblue}{rgb}{0.21,0.49,0.74}
\usepackage[pagebackref,breaklinks,colorlinks,allcolors=cvprblue]{hyperref}

\newcommand*\videogen{\textsc{Phantom}}

\NewDocumentCommand{\ying}{ mO{} }{\textcolor{teal}{\textsuperscript{\textit{Ying}}\textsf{\textbf{\small[#1]}}}}

\usepackage{cuted}
\usepackage{microtype}
\usepackage{amsmath,amsfonts,bm}

\usepackage{inconsolata}
\usepackage[capitalize]{cleveref}
\crefformat{equation}{Eq.~(#2#1#3)}
\crefname{section}{§}{§§}
\Crefname{section}{§}{§§}
\usepackage{booktabs}       
\usepackage{nicefrac}       
\usepackage{xcolor}         
\usepackage{graphicx}
\usepackage{multirow}
\usepackage{caption}
\usepackage{subcaption}
\usepackage{amssymb}
\usepackage{listings}

\usepackage{colortbl}
\usepackage{array}
\usepackage{environ}
\usepackage{makecell}
\usepackage{algorithm}
\usepackage{algpseudocode} 
\usepackage[most]{tcolorbox}
\usepackage{tikz}
\usepackage{mathtools}

\crefformat{equation}{Eq.~(#2#1#3)}
\crefname{section}{§}{§§}
\Crefname{section}{§}{§§}

\usepackage{newtxtext}

\definecolor{pgreen}{RGB}{34,139,34} 
\newcommand{\increase}[1]{$_{\mathbf{\textcolor{pgreen}{\uparrow #1}}}$}

\usepackage{amsmath,amsfonts,bm}

\def\eqref#1{equation~\ref{#1}}

\def\1{\bm{1}}

\def\vh{{\bm{h}}}

\def\vu{{\bm{u}}}
\def\vv{{\bm{v}}}

\def\vx{{\bm{x}}}

\def\vz{{\bm{z}}}

\def\mW{{\bm{W}}}

\DeclareMathAlphabet{\mathsfit}{\encodingdefault}{\sfdefault}{m}{sl}
\SetMathAlphabet{\mathsfit}{bold}{\encodingdefault}{\sfdefault}{bx}{n}

\newcommand{\E}{\mathbb{E}}

\def\paperID{} 
\def\confName{CVPR}
\def\confYear{2026}

\definecolor{IllinoisOrange}{HTML}{FF5F05}
\definecolor{IllinoisBlue}{HTML}{13294B}
\newcommand{\logoicon}{%
  \raisebox{-0.20\height}{\includegraphics[height=1.2em]{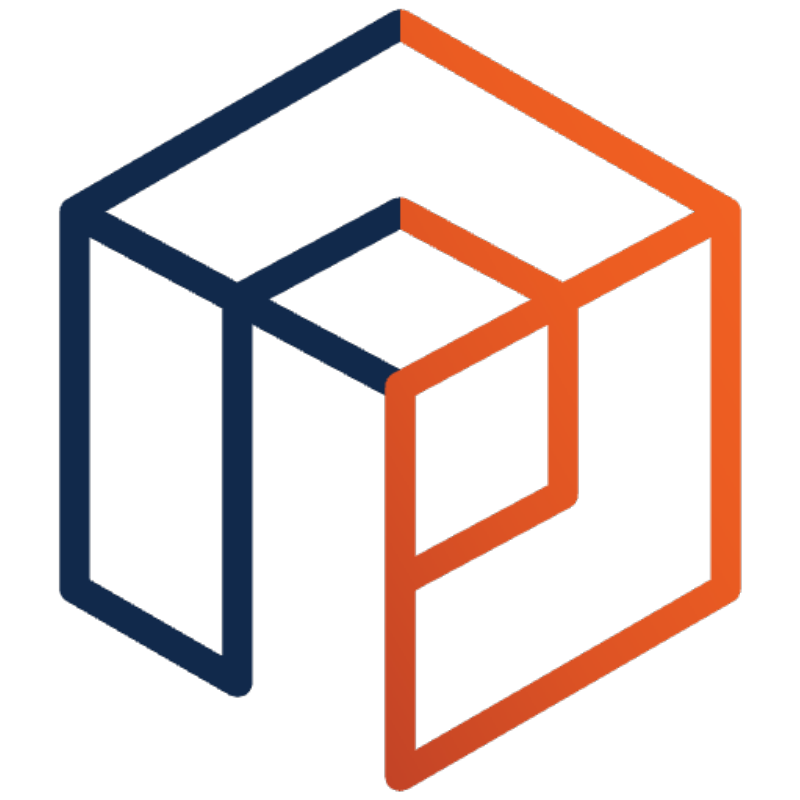}}%
}
\title{\videogen{}: \underline{Ph}ysics-Infused Video Gener\underline{a}tio\underline{n} via Join\underline{t} M\underline{o}deling of Visual and Latent Physical Dyna\underline{m}ics}

\author{Ying Shen \quad Jerry Xiong \quad Tianjiao Yu \quad Ismini Lourentzou\\
University of Illinois Urbana-Champaign\\
{\tt\small \{ying22,jerryx5,ty41,lourent2\}@illinois.edu}
}

\begin{document}
\maketitle

\begin{strip}
    \centering
    \vspace{-5em}
    \includegraphics[width=\linewidth]{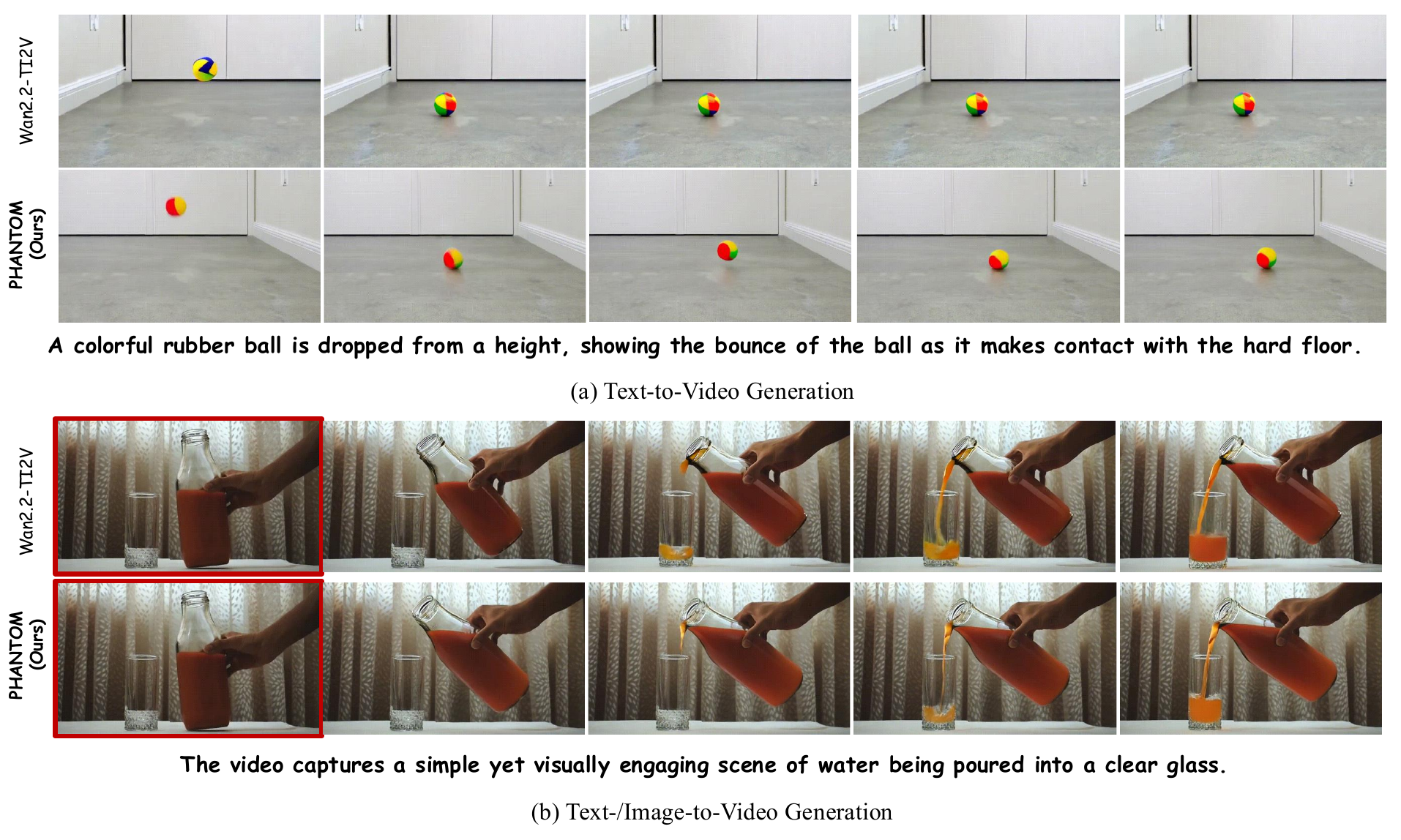}
    \vspace{-2.2em}
     \captionof{figure}{\textbf{Comparison between the base model Wan2.2-TI2V~\citep{wan2025} and our \videogen{} model in both (a) text-to-video and (b) text/image-to-video generation. 
     Red boxes mark the conditioning frame. }
     In (a), Wan2.2-TI2V fails to model the correct bouncing dynamics of the falling ball: after hitting the ground, the ball unnaturally loses all momentum and stops abruptly. In contrast, \videogen{} produces a physically plausible sequence in which the ball continues to bounce in accordance with real-world dynamics.
     In (b), when conditioned on an initial frame showing an empty glass, the juice suddenly appears at the bottom of the glass before any liquid is poured in Wan2.2-TI2V, while \videogen{} keeps the glass empty until liquid is actually poured in.
     }
    \label{fig:teaser}
\end{strip}

\begin{abstract}
Recent advances in generative video modeling, driven by large-scale datasets and powerful architectures, have yielded remarkable visual realism. However, emerging evidence suggests that simply scaling data and model size does not endow these systems with an understanding of the underlying physical laws that govern real-world dynamics. Existing approaches often fail to capture or enforce such physical consistency, resulting in unrealistic motion and dynamics.
In this work, we investigate whether integrating the inference of latent physical properties directly into the video generation process can equip models with the ability to produce physically plausible videos.
To this end, we propose \textbf{\videogen{}}, a \underline{Ph}ysics-Infused Video Gener\underline{a}tio\underline{n} model that join\underline{t}ly m\underline{o}dels the visual content and latent physical dyna\underline{m}ics.
Conditioned on observed video frames and inferred physical states, \videogen{} jointly predicts latent physical dynamics and generates future video frames.
\videogen{} leverages a physics-aware video representation that serves as an abstract yet informative embedding of the underlying physics, facilitating the joint prediction of physical dynamics alongside video content without requiring an explicit specification of a complex set of physical dynamics and properties.
By integrating the inference of physical-aware video representation directly into the video generation process, \videogen{} produces video sequences that are both visually realistic and physically consistent.
Quantitative and qualitative results on both standard video generation and physics-aware benchmarks demonstrate that \videogen{} not only outperforms existing methods in terms of adherence to physical dynamics but also delivers competitive perceptual fidelity.

\noindent \logoicon~\href{https://plan-lab.github.io/phantom}{\textcolor{IllinoisBlue}{PLAN Lab}~\textcolor{IllinoisOrange}{https://plan-lab.github.io/phantom}}

\end{abstract}

\section{Introduction}
Generative video modeling~\citep{OpenAI2024Sora,DeepMind2024Veo2,MetaAI2024MovieGen,susladkar2026pyratok} has advanced rapidly in recent years, driven by large-scale datasets and increasingly powerful generative architectures~\citep{vaswani2017attention,ho2020denoising,ho2022video,chenflow,peebles2023scalable}. 
These advancements have enabled impressive video synthesis capabilities, producing high-fidelity, visually plausible, and even surreal video sequences.
As these models become more capable, there is growing interest in whether generative video models can evolve into a form of world models~\citep{ha2018recurrent,lecun2022path,OpenAI2024Sora,shen2026egoforge}, systems that not only generate visually plausible video frames but also develop an intrinsic understanding of the fundamental laws of physics, ensuring that generated frames adhere to real-world principles.

Despite their visual fidelity, current video generation models continue to struggle with generating videos that comply with the fundamental physical principles that govern real-world dynamics~\citep{meng2024towards,kang2024far,motamed2025generative}. This disconnect highlights a gap between visually plausible video synthesis and true physical understanding, and raises a fundamental question: \emph{Can generative video models grasp the physical principles that govern reality simply by scaling up training on ever-larger video datasets with a next-frame prediction objective?}

Recent work~\citep{kang2024far} suggests that simply scaling model capacity or dataset size is insufficient for learning generalizable physical laws. Instead of abstracting general physical rules, models appear to rely on memorization, exhibiting case-based imitation for out-of-distribution generalization rather than internalizing fundamental principles.
We hypothesize that the inability of current video generation models to learn physical dynamics stems from their predominant reliance on the next-frame prediction objective. 
While effective for generating visually plausible content, this objective does not explicitly enforce physical reasoning, making it difficult for models to internalize and adhere to real-world physical laws.
To overcome this limitation, we argue that video generative models should jointly model the prediction of video content and latent physical parameters.\looseness-1

To this end, we introduce \textbf{\videogen{}}, a \underline{Ph}ysics-Infused Video Gener\underline{a}tio\underline{n} model that join\underline{t}ly m\underline{o}dels the visual content and latent physical dyna\underline{m}ics.
\videogen{} explicitly incorporates physical reasoning into the video generative process by augmenting a pretrained video diffusion model with a dedicated physical dynamics branch.
This physics branch is trained to infer and predict latent physical dynamics alongside video content, conditioned on both observed frames and current physical states.
Specifically, we leverage the latent embedding space of V-JEPA2~\citep{assran2025v}, a pretrained vision encoder shown to capture video representations that achieve an understanding of various intuitive physics properties~\citep{garrido2025intuitive}.
These embeddings serve as the latent representation of the underlying physics, enabling the model to reason about physical interactions and behaviors without requiring explicit specification of physical properties, simulator access, or external test-time reasoning.
Using the pretrained physics-aware embeddings extracted from observed video frames as latent physical representations, \videogen{} is trained to jointly predict future frames and their corresponding physics-aware embeddings, conditioned on both current visual content and associated latent physical states.\looseness-1

By integrating latent physical dynamics directly into the video generation pipeline, our approach encourages the model not only to generate visually plausible video sequences but also to understand how physical parameters evolve over time.
Quantitative evaluations on both standard video generation and physics-aware benchmarks demonstrate that our method significantly improves physical consistency without sacrificing visual realism.
Across three comprehensive physics-focused benchmarks, \videogen{} consistently outperforms the base model Wan2.2-TI2V~\citep{wan2025}, achieving a 50.4\% PC improvement on VideoPhy, a 2.6\% PC improvement on VideoPhy-2, and a 33.9\% gain on Physics-IQ.

\noindent The contributions of this work are as follows:
\begin{itemize}
\item We introduce \textbf{\videogen{}}, a physics-infused video generation framework that jointly models visual content and latent physical dynamics within a unified generative process.
\item Rather than relying on external simulators or inference-time guidance, we propose a dual-branch flow-matching architecture that couples a pretrained video generator with a dedicated physics branch operating in a physics-aware latent space, enabling the model to infer, evolve, and exchange physical state information during generation through bidirectional cross-attention.
\item We demonstrate the effectiveness of \videogen{} in producing video sequences that are both perceptually realistic and physically coherent through extensive experiments on standard video generation and physics-aware benchmarks.
\end{itemize}

\section{Related Work}

\noindent \textbf{Video Diffusion Models and Flow Matching.}
With the growth of large-scale video data and generative architectures, video diffusion models have achieved remarkable success.
Diffusion probabilistic models~\citep{ho2020denoising,songdenoising,songscore} and flow-matching models~\citep{lipmanflow,chenflow,liuflow} have emerged as powerful paradigms for modeling high-dimensional visual data, enabling high-fidelity generation across both images and videos.
Building on this foundation, large-scale text-to-video diffusion models such as Sora~\citep{OpenAI2024Sora}, HunyuanVideo~\citep{kong2024hunyuanvideo}, and Wan2.2-TI2V-5B~\citep{wan2025} have demonstrated impressive visual realism, temporal coherence, and open-domain generalization.
These models demonstrate the effectiveness of scaling diffusion-based architectures for complex video generation tasks, but remain primarily optimized for visual fidelity rather than physical correctness.
More recent work incorporates auxiliary signals or richer learned representations into generative modeling~\citep{kouzelisboosting,chefer2025videojam,yurepresentation}. REPA~\citep{yurepresentation} aligns denoising features with pretrained visual representations, ReDi~\citep{kouzelisboosting} jointly models visual latents and PCA-compressed high-level semantic features for image generation, and VideoJAM~\citep{chefer2025videojam} introduces optical-flow supervision to learn joint appearance-motion representations for video.
In contrast, our work targets physical-aware video generation and adopts a dual-branch architecture that freezes the visual backbone and trains a lightweight physics branch, enabling physics-aware reasoning at lower cost while preserving visual quality.

\noindent \textbf{Physics-aware Video Generation.}
While modern video generation models excel at visual synthesis, their physical plausibility remains limited, generating videos that often violate basic principles of motion, gravity, or material interaction~\citep{bansalvideophy,bansal2025videophy,motamed2025generative,kang2024far}. To address these shortcomings, several research directions have emerged.
One line of work integrates physical simulators or differentiable physics engines into the generative pipeline.
Works such as PhysAnimator~\citep{xie2025physanimator}, PhysGen~\citep{liu2024physgen}, and MotionCraft~\citep{montanaro2024motioncraft}, leverage physics simulators to guide motion generation or constrain predicted trajectories.
While effective within the simulator’s domain, such approaches are limited by the fidelity, assumptions, and coverage of the underlying physics engines, hindering generalization to real-world scenarios.
Another line improves physical realism through prompt-level or inference-time guidance~\citep{xue2025phyt2v,zhang2025think,huang2025vchain}, using external knowledge or MLLM reasoning to refine prompts or intermediate generations. DiffPhy~\citep{zhang2025think} infers physical context from prompts to guide diffusion, while PhyT2V~\citep{xue2025phyt2v} refines prompts through multi-step reasoning.
Although such strategies enhance physical plausibility, they operate outside the video generative model, do not increase the model's intrinsic physical understanding, and introduce substantial inference overhead.
A complementary thread uses representation alignment to inject physical priors. VideoREPA~\citep{zhang2025videorepa}, for instance, aligns video diffusion model latents with self-supervised video model features to encourage more physically grounded dynamics. However, such alignment is indirect and does not explicitly model the evolution of physical states.\looseness-1
Different from these approaches, \videogen{} integrates physical reasoning \textit{directly} into the generative process by jointly modeling physics-aware latent embeddings alongside visual content, enabling the model to internalize and evolve physical dynamics during synthesis.

\section{Preliminaries}

\subsection{Flow Matching}

Flow-based generative models~\citep{lipmanflow,liuflow,albergobuilding} aim to learn a time-dependent velocity field $\vu^\theta_t$ 
that transports samples from a simple source distribution
 $p_0(\vx)$ (e.g., standard Gaussian) to a complex target distribution $p_1(\vx)$.
Recent work~\cite{lipmanflow,liuflow,albergobuilding} proposed a simple simulation-free Conditional Flow Matching (CFM) framework that directly regresses the velocity $\vu^\theta_t$ on a conditional vector field $\vu_t(\cdot \mid \vx_1)$:
\begin{align}
    \mathcal{L}_\theta = \E_{t, p_1(\vx_1), p_t(\vx_t \mid \vx_1)} \|\vu^\theta_t(\vx_t, t) - \vu_t(\vx_t \mid \vx_1)  \|^2,
\end{align}
where $p_t(\vx_t \mid \vx_1)$ defines the conditional probability paths over time $t \in [0, 1]$. Typically, we leverage a linear conditional flow that defines  $\vx_t = (1-t) \vx_1 + t \vx_0$ with the conditional velocity $\vu_t(\vx_t \mid \vx_1) = \vx_1 - \vx_0$.

At inference, we sample $x_0 \sim \mathcal{N}(0, 1)$ and compute $x_1 \sim p_1(x)$ by integrating the predicted velocity $\mathbf{u}_\theta(x_t, t)$ through the Ordinary Differential Equation (ODE) solver:
\begin{align}
    \frac{\mathrm{d} \vx_t}{\mathrm{d} t} = \vu^\theta_t (\vx_t).
    \label{eq:ode}
\end{align}

\section{\videogen{} Method}

\subsection{Problem Definition}

In this work, we study the problem of \textit{physics-infused joint video and physical dynamics generation}, where the objective is to jointly synthesize future video frames as well as latent physical dynamics.
Let $\vx^o = [x_1, x_2, \dots, x_{t}]$ denote a sequence of observed video frames, and let $c$ be an optional textual prompt that provides contextual or semantic information about the scene. 
The goal is to predict a sequence of future video frames $\vx^f = [x_{t+1}, \dots, x^{T}]$ along with the corresponding latent physical dynamics $\vz^f$, conditioned on the observed video frames $\vx^o$ and physical dynamics $\vz^o$.

This task can be formulated as modeling the joint conditional distribution:
\begin{equation}
    p_\theta(\vx^f, \vz^f \mid \vx^o, \vz^o, c),
\end{equation}
where $\theta$ denotes the model parameters. 
The latent physical states $\vz^o$ capture physically meaningful properties encoded in a learned physics-aware representation space. 

The central motivation behind this formulation is to endow video generative models with an internal understanding of dynamics: rather than predicting future pixels solely from appearance cues, \videogen{} jointly infers and evolves latent physical states alongside visual content. This joint modeling encourages the resulting video generations to be not only visually plausible but also consistent with physical principles governing real-world dynamics.

\begin{figure*}[!t]
    \centering
    \includegraphics[width=0.95\linewidth]{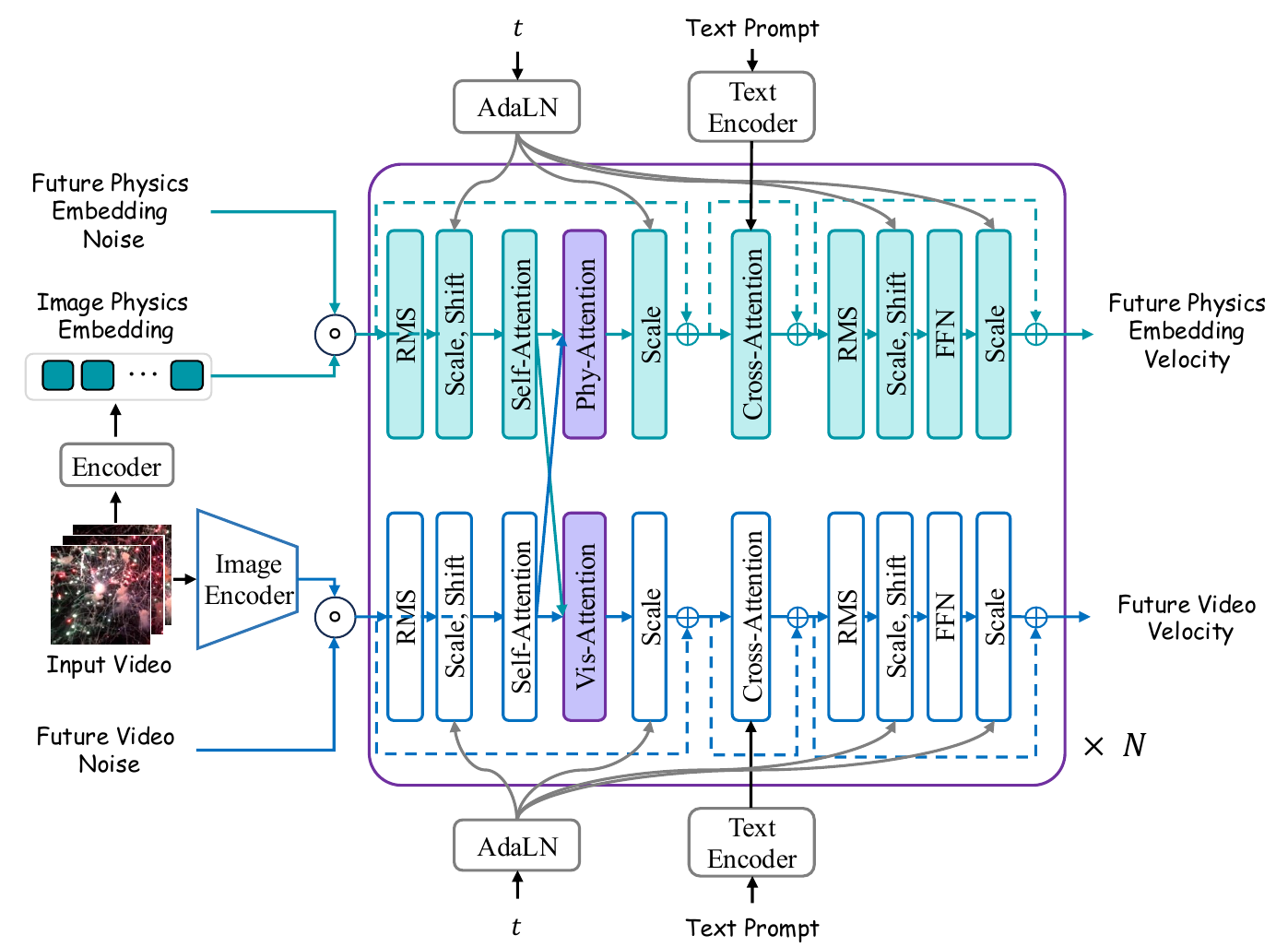}
    \vspace{-0.2cm}
    \caption{\textbf{\videogen{} Overview.} \videogen{} consists of two parallel latent flow-matching branches: the video branch and physics branch. 
    These branches jointly model future visual and physical dynamics, \ie, the video branch (white) predicts future visual trajectories, while the physics branch (teal) predicts the evolution of latent physical states. 
    Dual cross-attention layers tightly couple these branches, allowing physics cues to guide visual generation and visual evidence to refine physics reasoning.
     Color-filled components indicate trainable modules within the architecture. This design equips \videogen{} with an internal model of dynamics, enabling physically consistent video prediction
    }
    \label{fig:physgen}
\end{figure*}

\subsection{Physics-Infused Video Generation}
\label{sec:method}
To address the task of joint video and physical-dynamics generation, we propose \textbf{\videogen{}}, a \underline{Ph}ysics-Infused Video Gener\underline{a}tio\underline{n} model that join\underline{t}ly m\underline{o}dels the visual content and latent physical dyna\underline{m}ics.
Specifically, 
\videogen{} adopts a dual-branch architecture that simultaneously predicts future video frames and their corresponding latent physical states.
Built on top of Wan2.2-TI2V~\citep{wan2025}, a pretrained latent video diffusion model that supports text-to-video and text-/image-to-video generation, \videogen{} augments the visual generation pathway with a parallel physical dynamics branch that enables the model to explicitly reason over latent physical processes inferred from observed video sequences.
An overview of the architecture is shown in \Cref{fig:physgen}.

Given an observed video sequence $\vx^o$, we first encode it into two complementary latent spaces: (1) a visual latent sequence $\vv^o$ representing low-level visual appearance, and (2) a physical latent sequence $\vz^o$ representing high-level, inferred physical dynamics. 
The visual representation $\vv^o$ is obtained via a pretrained video VAE encoder $\mathcal{E}_v$, such that $\vv^o=\mathcal{E}_v(\vx^o)$. 
Simultaneously, the latent physical state $\vz^o$ is derived using V-JEPA2~\citep{assran2025v}, a self-supervised video encoder, producing $\vz^o=\mathcal{E}_{\text{V-JEPA2}}(\vx^o)$. 
Prior work~\citep{garrido2025intuitive} has shown that V-JEPA2's representations encode a strong understanding of intuitive physics concepts, such as object permanence, collisions, and gravity, making it a suitable representation for underlying physical dynamics.

\videogen{} consists of two parallel latent flow-matching branches. 
The video branch reuses the pretrained Wan2.2~\citep{wan2025} modules to process the visual latent sequence, while the physics branch mirrors the architecture and is adapted to predict physical dynamics in the latent space. 
Although each branch maintains its own modality-specific hidden states, they exchange information through two cross-attention layers inserted at corresponding depths in both branches. Specifically, the Vis-Attention module in the video branch attends to the hidden states of the physics branch, while the Phy-Attention module symmetrically attends to the hidden states of the video branch, as illustrated in \Cref{fig:physgen}. This design enables the model to coordinate visual and physical reasoning while preserving the expressive capacity of each modality.\looseness-1

Concretely, the Vis-Attention and Phy-Attention modules compute the updated hidden states for the two branches as follows:
\begin{align}
    \vh_v' = \textrm{Softmax}(\frac{(\mW^Q_v \vh_v) (\mW^K_v \vh_{z})^T}{\sqrt{d}}) (\mW^V_v \vh_{z}) \\
    \vh_z' = \textrm{Softmax}(\frac{(\mW^Q_z \vh_z) (\mW^K_z \vh_{v})^T}{\sqrt{d}}) (\mW^V_z \vh_{v}),
\end{align}
where $\mW^Q_v, \mW^K_v, \mW^V_v$ and $\mW^Q_z, \mW^K_z, \mW^V_z$ are the learnable projection matrices for the video and physics branches, respectively, and $d$ is the latent feature dimension.

This dual-cross attention design allows each branch to maintain modality-specific representations while enabling dynamic information exchange between two branches, without collapsing the two modalities into a single entangled representation. In practice, dual-cross attention provides finer control than joint-attention alternatives and avoids the instability and undesired feature entanglement observed when visual and physical states are mixed too aggressively.

Through this cross-modal coupling, \videogen{} learns rich correspondences between visual and physical dynamics, which are essential for generating sequences that are both visually coherent and physically consistent. Conditioning signals, including the textual prompt $c$ and the flow-matching timestep $t$, are injected into both branches to ensure aligned conditioning throughout the generation process.

\vspace{0.1cm}
\noindent \textbf{Training Strategies.}
During training, we freeze all pretrained parameters in the video branch to preserve its strong generative prior, and update only the physics branch together with the dual cross-attention layers. The trainable components are highlighted in color in \Cref{fig:physgen}. This selective adaptation strategy enables the model to incorporate physical reasoning while preserving the visual generation quality of the pretrained backbone.

To enable \videogen{} to operate in a video-to-video setting,  we extend Wan2.2-TI2V beyond its native text- or single-image-conditioning setup to accept an arbitrary number of conditioning frames during training. 
Following Wan2.2's design, these conditioning frames are concatenated with the noised future frames along the temporal dimension, with their flow-matching timestep being fixed to $t\!=\!0$, ensuring that these frames remain unperturbed and provide deterministic conditioning inputs for predicting future dynamics.

We adopt the standard flow-matching objective~\citep{chenflow}, extending it to jointly learning the target velocity field of both video $\vu_t(\vv_t^f \mid \vv^f_1)$ and physical dynamics $\vu_t(\vz_t^f \mid \vz^f_1)$ at timestep $t$.
The training loss is defined as: 
\begin{equation}
\scalebox{0.97}{$
\begin{aligned} 
&\mathcal{L}(\theta) = \mathbb{E}_{t,p_1(\vv_1^f), p_1(\vz_1^f), p_t(\vv_t^f|\vv_1^f), p_t(\vz_t^f|\vz_1^f)} 
 \\
 &\left\| \vu^\theta_t(\vv^f_t, \vz^f_t, \vv^o_1, \vz^o_1, t, c) - [\vu_t(\vv_t^f |\vv^f_1) ; \vu_t(\vz_t^f |\vz^f_1)] \right\|^2 ,
\end{aligned}     
 $}
\end{equation}
where $p_0(\cdot)$ and $p_1(\cdot)$ are the source and target endpoint distributions in the flow-matching framework.  
For clarity, we decompose the loss into visual and physical components by extracting the corresponding predicted velocity from the joint model output:
\begin{align}
    \mathcal{L}_v &= \left\| \vu^\theta_t(\vv^f_t, \vz^f_t, \vv^o_1, \vz^o_1, t, c)[\vv] - \vu_t(\vv_t^f |\vv^f_1)  \right\|^2  \\
    \mathcal{L}_z &= \left\| \vu^\theta_t(\vv^f_t, \vz^f_t, \vv^o_1, \vz^o_1, t, c)[\vz] - \vu_t(\vz_t^f |\vz^f_1) \right\|^2  \\
    \mathcal{L}(\theta) &= \mathcal{L}_v + \alpha_z \mathcal{L}_z,
\end{align}
where $\vu^\theta_t(\cdot)[\vv]$ and $\vu^\theta_t(\cdot)[\vz]$ denote the visual and physical components of the predicted velocity, respectively, and $\alpha_z$ controls the contribution of the physical loss $\mathcal{L}_z$.

In practice, we observe that the magnitude and gradient norm of physical loss $\mathcal{L}_z$ is substantially larger than that of visual loss $\mathcal{L}_v$, which can destabilize training.
To address this issue, we employ a \textit{recursive loss-weight scheduling strategy}.
Specifically, we initialize $\alpha_z\!=\!0$ and gradually increase it over training. 
Once the gradient norm of the physics branch exceeds a predefined threshold $\eta_z$, we reset  $\alpha_z$ back to zero and restart the schedule. This cyclic weighting stabilizes optimization by preventing the physics branch from overwhelming the shared architecture while still allowing it to contribute meaningful gradients over time. 
Through joint optimization, \videogen{} produces video sequences that are not only visually realistic but also more consistent with the underlying physical dynamics of the scene.

\section{Experimental Setup}

\noindent \textbf{Datasets.}
We train \videogen{} on OpenVidHD-0.4M~\citep{nan2024openvid}, a high-quality subset of the OpenVid-1M dataset containing approximately 400K high-resolution video–text pairs. Importantly, this dataset provides diverse visual content but is not explicitly designed to emphasize physical dynamics.

\noindent \textbf{Evaluation.}
We employ a suite of complementary benchmarks to evaluate both the general generative quality and physical awareness of \videogen{}.
\begin{itemize}
    \item \textbf{General Generative Quality.}
We assess overall video generation capability using \textbf{VBench-2}~\citep{zheng2025vbench}, a structured and widely adopted benchmark designed to measure the intrinsic faithfulness of generative video models. VBench-2 evaluates five core dimensions, Human Fidelity, Controllability, Creativity, Physics, and Commonsense, across 18 fine-grained metrics, providing a comprehensive assessment of overall video quality.
\item \textbf{Physics-Focused Evaluation.}
To specifically assess the physical plausibility of generated videos, we further evaluate on \textbf{VideoPhy}~\citep{bansalvideophy}, \textbf{VideoPhy2}~\citep{bansal2025videophy}, and \textbf{Physics-IQ}~\citep{motamed2025generative}.
To specifically assess physical plausibility, we further evaluate on VideoPhy~\citep{bansalvideophy}, VideoPhy2~\citep{bansal2025videophy}, and Physics-IQ~\citep{motamed2025generative}. VideoPhy focuses on semantic adherence and physical commonsense across diverse material types and interactions. VideoPhy2 extends this benchmark with an action-centric design that incorporates human interactions, serving as a larger, more complex, and more rigorous version.
Physics-IQ provides a real-world benchmark featuring both single-frame and multi-frame evaluation settings with real-world reference videos, enabling detailed assessment of physical plausibility and reasoning consistency across diverse physical phenomena.
\end{itemize}

\noindent \textbf{Baselines.}
We compare \videogen{} against both state-of-the-art video generation models, including CogvideoX~\citep{yang2024cogvideox}, HunyuanVideo~\citep{kong2024hunyuanvideo}, Wan2.2-TI2V~\citep{wan2025}.
To further assess physics awareness, we also include physics-aware methods like PhyT2V~\citep{xue2025phyt2v}, VideoREPA~\citep{zhang2025videorepa}, and WISA~\citep{wang2025wisa}. 
This set of baselines enables us to evaluate \videogen{} against strong generative models in terms of overall video quality, while also assessing its ability to improve physical realism.

\noindent \textbf{Implementation Details.}
We build upon Wan2.2-TI2V-5B~\citep{wan2025}, a powerful text–image-to-video diffusion model, and extend it with a dual-branch architecture that jointly models visual content and latent physical dynamics. The physics branch is initialized from scratch, while all pretrained visual-branch parameters remain frozen during training to preserve Wan’s strong generative prior. 
We further extend the base architecture to support multi-frame conditioning, enabling the model to process up to 121 frames at a resolution of 480 $\times$ 832.
During training, the number of conditioning frames is randomly sampled between 1 and 45 to expose the model to varying temporal context lengths in text-/video-to-video mode, while in 50\% of training instances, no conditioning frames are provided, corresponding to text-to-video generation.\looseness-1

\subsection{Quantitative Results}
We evaluate \videogen{} across multiple physics-aware video generation benchmarks to assess both general visual quality and physical consistency.
We first evaluate \videogen{}'s text-to-video generation performance on VideoPhy~\citep{bansalvideophy} and VideoPhy-2~\citep{bansal2025videophy}, two physics-based benchmarks focused on physical commonsense and action-conditioned physical reasoning. 
For both benchmarks, we adopt their official auto-evaluators to compute the Physical Commonsense (PC) and Semantic Adherence (SA) metrics.

As reported in Table~\ref{table:videophy}, \videogen{} delivers consistent gains over its pretrained Wan2.2-TI2V backbone across both benchmarks, validating the benefit of explicitly modeling latent physical dynamics. On VideoPhy, \videogen{} improves semantic adherence by 14.5\% and physical commonsense by 50.4\%, achieving the best PC score (37.9) among all compared methods. On VideoPhy-2, \videogen{} also demonstrates a notable gain of 13.1\% on SA score and 2.6\% on PC score over the baseline, further validating its ability to capture intricate physical dynamics.
\begin{table}[!t]
  \centering
    \caption{\textbf{VideoPhy and VideoPhy2 Results.} Semantic Adherence (SA) measures video-text alignment and fidelity. Physical Commensense (PC) measures whether generated videos follow real-world physics laws intuitively. $\dagger$ denotes the results reported from VideoREPA~\citep{zhang2025videorepa} with the original prompt input. Improvements over the base model Wan2.2-TI2V are highlighted in \increase{green}. Best results shown in \textbf{bold}, second-best \underline{underlined}. 
  }
  \vspace{-0.3cm}
  \resizebox{\linewidth}{!}{%
  \begin{tabular}{p{4cm} p{1.8cm} p{1.8cm} p{1.8cm} p{1.8cm}}
    \toprule
    \textbf{Method} & \multicolumn{2}{c}{\textbf{VideoPhy}} & \multicolumn{2}{c}{\textbf{VideoPhy-2}} \\ \cmidrule(lr){2-3} \cmidrule(lr){4-5} 
    & \textbf{SA}$\uparrow$ & \textbf{PC}$\uparrow$  &  \textbf{SA}$\uparrow$ & \textbf{PC}$\uparrow$  \\ 
    \midrule
    \rowcolor{gray!15} \multicolumn{5}{l}{\textit{General-Purpose}} \\ 
    VideoCrafter2~\citep{chen2024videocrafter2} & 50.3 & 29.7 & 25.89 & 55.67 \\
    LaVIE~\citep{wang2025lavie} & 48.7 & 31.5 & - & - \\
    Cosmos-Diffusion-7B~\citep{agarwal2025cosmos} & \underline{57.0} & 18.0 & 26.32 & 54.19 \\
    CogVideoX-5B~\citep{yang2024cogvideox} & \textbf{63.1} & 31.4 & \textbf{28.86} & 68.42 \\
    Wan2.2-TI2V-5B~\citep{wan2025} & 41.5 & 25.2 & 24.53 & 69.20 \\ 
    \rowcolor{gray!15} \multicolumn{5}{l}{\textit{Physics-Focused}} \\ 
    PhyT2V (Round 4)$\dagger$~\citep{xue2025phyt2v} & 61 & \underline{37} & - & -\\ 
    WISA$\dagger$~\citep{wang2025wisa} & 62 & 33 & - & -\\ 
    VideoREPA~\citep{zhang2025videorepa} & 51.9  & 22.4  & 21.02 & \textbf{72.54} \\
    \rowcolor{blue!15}\videogen{}$_{\textrm{Wan2.2}}$ (Ours) &  47.5\increase{14.5\%} & \textbf{37.9}\increase{50.4\%} & \underline{27.75}\increase{13.1\%} & \underline{71.74}\increase{2.6\%}\\
    \bottomrule
  \end{tabular}
  }
  \label{table:videophy}
  \vspace{-0.3cm}
  \end{table}

\begin{table*}[!t]
  \centering
    \caption{\textbf{Physics-IQ Results.} Baselines missing from the multi-frame setting do not support multi-frame conditioning. Improvements over the base model Wan2.2-TI2V are highlighted in \increase{green}. The best results are shown in \textbf{bold}, and the second-best are \underline{underlined}.}
    \vspace{-0.3cm}
  \resizebox{\linewidth}{!}{%
  \begin{tabular}{l  l   c c c c  c}
  \toprule
     & \textbf{Method}  & \textbf{Spatial IoU $\uparrow$} & \textbf{Spatiotemporal IoU $\uparrow$} & \textbf{Weighted spatial IoU $\uparrow$} & \textbf{MSE $\downarrow$} & \textbf{Physics-IQ Score $\uparrow$}\\
  \midrule

\multirow{8}{*}{\makecell[c]{Single\\Frame}}
  & \multicolumn{6}{>{\columncolor{gray!12}}l}{\textit{General-Purpose}}\\
  & VideoPoet~\citep{kondratyuk2024videopoet} & 0.141 & 0.126 & 0.087 & 0.012 & 20.30 \\
  & Lumiere~\citep{bar2024lumiere}            & 0.113 & \underline{0.173} & 0.061 & 0.016 & 19.00 \\
  & Runway Gen 3~\citep{runway2024}           & \underline{0.201} & 0.115 & 0.116 & 0.015 & 22.80 \\
  & CogVideoX1.5-I2V~\citep{yang2024cogvideox}& 0.198 & \textbf{0.189} & \underline{0.127} & 0.015 & \underline{27.90} \\
  & Wan2.2-TI2V-5B~\citep{wan2025}            & 0.164 & 0.132 & 0.102 & \underline{0.010} & 22.10 \\
  & \multicolumn{6}{>{\columncolor{gray!12}}l}{\textit{Physics-Focused}}\\
  & RDPO~\citep{qian2025rdpo}                 &    -   &   -    &    -   &    -   &   25.21    \\
  & \cellcolor{blue!10} \textsc{\videogen{}} (Ours)               & \cellcolor{blue!10} ~~~~~~~~~~\textbf{0.245}\increase{49.4\%} & \cellcolor{blue!10} ~~~~~~~~~~0.146\increase{10.6\%} & \cellcolor{blue!10} ~~~~~~~~~~\textbf{0.140}\increase{37.3\%} & \cellcolor{blue!10} ~~~~~~~~~~\textbf{0.009}\increase{11.1\%} &  \cellcolor{blue!10}~~~~~~~~~~\textbf{29.59}\increase{33.9\%} \\
\midrule

  \multirow{4}{*}{\makecell[c]{\cellcolor{white}Multi-\\Frame}}
  & \multicolumn{6}{>{\columncolor{gray!12}}l}{\textit{General-Purpose}}\\
  & VideoPoet~\citep{kondratyuk2024videopoet} & \underline{0.204} & \textbf{0.164} & \textbf{0.137} & \textbf{0.010} & \textbf{29.50} \\
  & Lumiere~\citep{bar2024lumiere}            & 0.170 & \underline{0.155} & 0.093 & 0.013 & 23.00 \\
  \cline{2-7}
  & \multicolumn{6}{>{\columncolor{gray!12}}l}{\textit{Physics-Focused}}\\
   & \cellcolor{blue!10} \textsc{\videogen{}} (Ours)           &   \cellcolor{blue!10}\textbf{0.235} & \cellcolor{blue!10} 0.133 & \cellcolor{blue!10} \underline{0.132} & \cellcolor{blue!10} \underline{0.011} & \cellcolor{blue!10} \underline{27.53}  \\
  \bottomrule
  \end{tabular}}
  \label{table:physical_iq}
\end{table*}

To further assess generalization, we evaluate \videogen{} on Physics-IQ~\citep{motamed2025generative} under both single-frame and multi-frame conditioning settings. Physics-IQ measures a model's ability to infer and extrapolate physical dynamics from real-world motion sequences. Given either a single initial frame or a short observed clip, the model must predict future frames, which are then compared against ground-truth sequences to assess its understanding of the underlying physical behavior.

As shown in Table~\ref{table:physical_iq}, \videogen{} achieves substantial improvements over the Wan2.2-TI2V baseline in both conditioning setups, increasing the Physics-IQ score by 33.9\% in the single-frame setting and delivering competitive performance in the multi-frame setting, even though the base Wan2.2-TI2V model was not trained to support multi-frame conditioning. These results highlight the effectiveness of explicitly modeling latent physical dynamics.

\newcolumntype{L}[1]{>{\raggedright\arraybackslash}m{#1}}
\newcolumntype{C}[1]{>{\centering\arraybackslash}m{#1}}

\begin{table*}[t]
\centering
\caption{\textbf{Text-to-video evaluation on \textbf{VBench-2}. } Best results in \textbf{bold}.  Improvements over base model Wan2.2-TI2V highlighted in \increase{green}.
}
\vspace{-0.3cm}
\resizebox{0.99\linewidth}{!}{
\begin{tabular}{L{4cm} C{2.5cm} C{2.5cm} C{2.5cm} C{2.5cm} C{2.5cm} C{2.5cm}}
\toprule
\textbf{Model}  & \textbf{Total} &\textbf{Creativity}& \textbf{Commonsense} &\textbf{Controllability} &\textbf{Human Fidelity} &\textbf{Physics} \\
\midrule
Wan2.2-TI2V-5B~\citep{wan2025} & 51.57 & \textbf{52.50} & 60.57 & 18.50 & 86.10 & 40.19  \\ 
\rowcolor{blue!10} \videogen{} (Ours) &  ~~~~~~~~~\textbf{51.84}\increase{0.5\%} & 45.51 & ~~~~~~~~~\textbf{61.43}\increase{1.4\%} & ~~~~~~~~~\textbf{20.23}\increase{9.4\%} & ~~~~~~~~~\textbf{88.39}\increase{2.7\%} & ~~~~~~~~~\textbf{43.61}\increase{6.0\%} \\
\bottomrule
\end{tabular}}
\label{tab:vbench2}
 \vspace{-0.2cm}
\end{table*}

We additionally assess both perceptual quality and physical realism using VBench-2~\citep{zheng2025vbench}, a comprehensive evaluation suite for text-to-video models covering creativity, commonsense, controllability, human fidelity, and physics plausibility. 
As shown in \Cref{tab:vbench2}, \videogen{} achieves improvements over Wan2.2-TI2V across nearly all dimensions, with particularly large gains in Human Fidelity and Physics. These results indicate that incorporating latent physical dynamics not only enhances physical consistency but also improves the overall realism and stability of generated videos.

While \videogen{} shows a modest drop in the aggregate Creativity score, which comprises both diversity and composition, a fine-grained analysis shows that \videogen{} improves on Composition from 40.35 to 45.07 (+11.7\%) relative to Wan2.2-TI2V, but exhibits a reduction in Diversity (64.67 to 45.95).
One plausible explanation is that less physically plausible videos include unrealistic variations, which may inadvertently inflate diversity metrics.
Additional results for fine-grained VBench-2 dimensions can be found in the Appendix.
Overall, \videogen{} achieves a Total score on VBench-2 that is on par with, and slightly higher than Wan2.2-TI2V, indicating that the improvements in physics and fidelity are achieved without sacrificing overall video generation quality.\looseness-1

Across all benchmarks, \videogen{} demonstrates strong improvements on physics-related metrics while preserving competitive visual quality and semantic alignment, indicating that the integration of latent physical reasoning meaningfully enhances the physical coherence of generated videos.\looseness-1

\subsection{Qualitative Results}

\begin{figure*}[!t]
    \centering
    \includegraphics[width=0.99\linewidth]{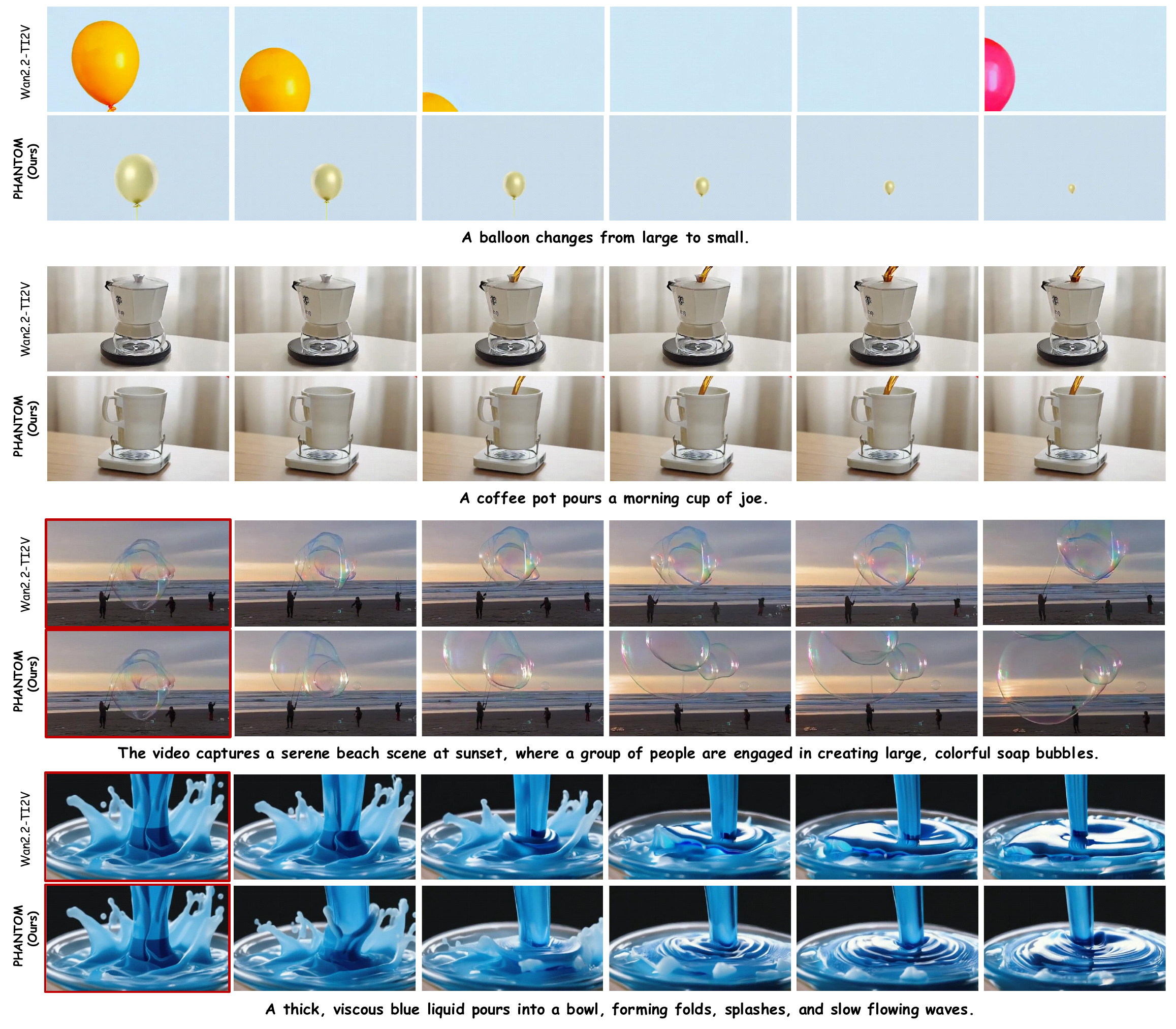}
    \vspace{-0.2cm}
    \caption{\textbf{Qualitative comparison between Wan2.2-TI2V~\citep{wan2025} and our \videogen{} across diverse text-to-video and text/image-to-video scenarios.} Red boxes indicate the conditioning frames. For prompts involving diverse physical processes, such as deformation, pouring, buoyancy, and viscous flow, \videogen{} produces motion that matches the requested behavior, while Wan2.2-TI2V often fails to follow the prompt or violates basic physical dynamics.
    Additional qualitative results are provided in the Appendix.}
    \label{fig:qualitative}
\end{figure*}

In \Cref{fig:qualitative}, we present qualitative comparisons to illustrate how \videogen{} improves physical plausibility and semantic consistency over the Wan2.2-TI2V baseline. Across diverse scenarios, including object deformation, pouring, buoyant motion, and viscous flow, \videogen{} generates dynamics that better match the intended physical process, whereas the baseline often exhibits semantic drift or implausible motion.

In the first example, the prompt describes a balloon \textit{changing from large to small}. Wan2.2-TI2V fails to realize this transformation: rather than shrinking the balloon, it effectively moves the balloon farther from the camera and even changes its color to red toward the end, violating both the described transformation and object identity. In contrast, \videogen{} correctly captures the intended physical transformation by generating a gradual, physically consistent shrinkage in balloon size while preserving identity and appearance.

In the second example, involving a \textit{coffee pot pouring into a mug}, Wan2.2-TI2V generates a mug with a lid, undermining the realism of the pouring action. The model proceeds to pour coffee as if the lid does not exist, resulting in an inconsistent and unrealistic sequence. In contrast, \videogen{} produces a lid-free mug and a more coherent pouring motion sequence that better aligns with real-world physical behavior.

We also evaluate challenging scenarios shown in Figure~\ref{fig:teaser}, where \videogen{} demonstrates more realistic interactions, such as proper bouncing, contact, and momentum transfer, compared to the baseline, which often causes objects to halt abruptly or follow implausible trajectories.
Notably, for the text-to-video samples, \videogen{} jointly denoises both the visual and physical latent spaces starting from pure noise, without requiring any externally provided physics-aware representation at inference time. This indicates that the model has internalized a latent understanding of physical behavior through joint training.

We also show text-/image-to-video examples (last two rows in Figure~\ref{fig:teaser}).
In the first example, which depicts \textit{people creating large soap bubbles on a beach at sunset}, Wan2.2-TI2V generates bubbles that behave more like rigid or semi-rigid objects, drifting with little meaningful deformation. In contrast, \videogen{} better captures the lightweight, deformable nature of soap bubbles: the produced bubbles stretch, wobble, and drift more naturally in the wind, better reflecting real-world physics and the softness of the material.

The last example shows a \textit{thick, viscous blue liquid pouring into a bowl}. In the later frames, Wan2.2-TI2V breaks physical realism, making the liquid appear to fall into an indefinite void rather than forming layered folds. \videogen{} produces a more physically coherent sequence, capturing the gradual buildup of fluid layers, the formation of folds, and the slow, flowing waves characteristic of high-viscosity liquids.

Across all qualitative examples, \videogen{} consistently demonstrates stronger alignment with the textual descriptions and improved adherence to underlying physical principles compared to Wan2.2-TI2V, highlighting the effectiveness of our proposed model.
Additional qualitative results and comparisons with existing methods are provided in the Appendix.\looseness-1

\section{Conclusion}
In this work, we introduce \textbf{\videogen{}}, a physics-infused video generation framework that jointly models visual content and latent physical dynamics. By coupling a pretrained video diffusion backbone with a dedicated physics-reasoning branch, \videogen{} learns to generate videos that respect both visual fidelity and intuitive physical laws, without relying on external simulators, prompt refinement, or post-hoc alignment.
Through extensive evaluation across physics-aware and general benchmarks, we demonstrate \videogen{} delivers substantial improvements in physical plausibility while preserving or enhancing perceptual quality. Qualitative results further demonstrate that \videogen{} produces sequences that respect momentum, collisions, fluid behavior, and material deformation, achieving competitive performance in both text-to-video and text-/image-to-video settings.\looseness-1

\section*{Acknowledgments}
This research was partially supported by Google, the Google TPU Research Cloud (TRC) program, the U.S. Defense Advanced Research Projects Agency (DARPA) under award HR001125C0303, and the U.S. Army under contract W5170125CA160. The views and conclusions contained herein are those of the authors and should not be interpreted as necessarily representing the official policies, either expressed or implied, of Google, DARPA, the U.S. Army, or the U.S. Government. The U.S. Government is authorized to reproduce and distribute reprints for governmental purposes notwithstanding any copyright annotation therein. This work also used the Delta GPUs provided by the National Center for Supercomputing Applications through allocations CIS250318 and CIS251228 from the Advanced Cyberinfrastructure Coordination Ecosystem: Services \& Support (ACCESS, \citet{boerner2023access}) program, supported by National Science Foundation grants \#2138259, \#2138286, \#2138307, \#2137603, and \#2138296.\\

{
    \small
    \bibliographystyle{ieeenat_fullname}
    \bibliography{main}
}

\appendix
\clearpage
\setcounter{page}{1}
\maketitlesupplementary

\section{Implementation Details}

\noindent \textbf{Training Details.}
For our main experiments, we build upon the Wan2.2-TI2V-5B~\citep{wan2025} due to its ability to accept both text and image inputs. We integrate our physics branch into this architecture as described in Section \ref{sec:method}. The physics branch is initialized from scratch, while the visual branch is kept frozen to preserve the strong generative prior of the base model.
For extracting physics-aware embeddings, we leverage V-JEPA2~\citep{assran2025v}, a pretrained video encoder shown to capture intuitive physics properties~\citep{garrido2025intuitive}. In particular, we use the VJEPA2-ViT-H-fpc64-256 variant. We have trained the model for two epochs.
We train all models with a global batch size of 128 using the AdamW optimizer with a learning rate of $4e-5$ and weight decay $1e-3$. We use cosine learning rate decay with a 5\% warmup ratio.
All experiments are performed on 4 NVIDIA H200 GPUs.

\noindent \textbf{Evaluation Details.}
We conduct evaluations on all benchmarks using their official protocols and codebases to ensure comparability with prior work.
For VideoPhy~\citep{bansalvideophy}, we use the official auto-rater for all evaluations. Results are reported using both the original prompts provided in the dataset and the more detailed prompts used in VideoREPA~\citep{zhang2025videorepa}. 
Following VideoREPA~\citep{zhang2025videorepa}, we set Semantic Adherence (SA) = 1 and Physical Commonsense (PC) = 1 when their values are greater than or equal to 0.5, and values less than 0.5 are set as SA = 0 and PC = 0. The final SA and PC scores correspond to the fraction of videos assigned a score of 1 after thresholding.\looseness-1

In VideoPhy-2~\citep{bansal2025videophy}, we follow the official evaluation protocol. Both SA and PC are computed as the proportion of videos that receive a rating of at least 4 out of 5 from the benchmark’s auto-evaluator. We directly use the official up-sampled prompts for evaluation.
For Vbench2~\citep{zheng2025vbench}, we report the results using its original prompts.

For Physics-IQ~\citep{motamed2025generative}, we evaluate under both single-frame and multi-frame conditioning. In the single-frame setting, the model receives only the initial frame and the caption as inputs, whereas in the multi-frame setting, the model observes a short initial clip and the corresponding caption.
 
\section{Baselines}

\subsection{General-Purpose Video Models}

We compare against several state-of-the-art general-purpose text-to-video (T2V) diffusion models that serve as strong baselines in open-domain video generation, including CogVideoX-5B~\citep{yang2024cogvideox}, HunyuanVideo~\citep{kong2024hunyuanvideo}, Wan2.1-T2I-14B, and Wan2.2-TI2V-5B~\citep{wan2025}. These models demonstrate strong open-domain generalization and high-fidelity video synthesis but are not designed to model or enforce physical principles.\looseness-1

\subsection{Physics-Focused Video Models}
In addition to general-purpose video generators, we compare against a set of recent physics-focused video generation approaches that aim to improve physical plausibility.

\noindent \textbf{PhyT2V~\citep{xue2025phyt2v}} uses large language models to iteratively refine prompts via chain-of-thought and step-back reasoning. By repeatedly analyzing and rewriting the prompt, it guides existing text-to-video models toward generating videos that better adhere to real-world physical laws without retraining the generation model.

\noindent \textbf{WISA~\citep{wang2025wisa}} is a physics-aware video generation approach that incorporates explicit physical categories and properties. These physical attributes are embedded into the generation process through Mixture-of-Physical-Experts Attention (MoPA) and a dedicated Physical Classifier, enabling the model to incorporate richer physical priors during synthesis.

\noindent \textbf{VideoREPA~\citep{zhang2025videorepa}} injects physics understanding into diffusion-based video generators by aligning their hidden states with the representation from video foundation models via distillation.

\section{Additional Results}

\begin{table}[!t]
  \centering
    \caption{\textbf{Results on VideoPhy and VideoPhy2 Benchmarks.} Semantic Adherence (SA) measures video-text alignment and fidelity. Physical Commensense (PC) measures whether generated videos follow real-world physics laws intuitively. $\dagger$ denotes results reported from VideoREPA~\citep{zhang2025videorepa} with the original prompt. Improvements over the base model Wan2.2-TI2V are highlighted in \increase{green}. Best results in \textbf{bold}, second-best \underline{underlined}. Following VideoREPA~\citep{zhang2025videorepa}, we also report results with detailed prompts, denoted by $^*$.
  }
  \vspace{-0.3cm}
  \resizebox{0.99\linewidth}{!}{%
  \begin{tabular}{p{3.5cm}  c c c c}
    \toprule
    \textbf{Method} & \multicolumn{2}{c}{\textbf{VideoPhy}} & \multicolumn{2}{c}{\textbf{VideoPhy-2}} \\ \cmidrule(lr){2-3} \cmidrule(lr){4-5} 
    & \textbf{SA}$\uparrow$ & \textbf{PC}$\uparrow$  &  \textbf{SA}$\uparrow$ & \textbf{PC}$\uparrow$  \\ 
    \midrule
    \rowcolor{gray!15} \multicolumn{5}{l}{\textit{General-Purpose}} \\ 
    VideoCrafter2~\citep{chen2024videocrafter2} & 50.3 & 29.7 & 25.89 & 55.67 \\
    LaVIE~\citep{wang2025lavie} & 48.7 & 31.5 & - & - \\
    Cosmos-Diffusion-7B~\citep{agarwal2025cosmos} & 57.0 & 18.0 & 26.32 & 54.19 \\
    CogVideoX-5B~\citep{yang2024cogvideox} & 63.1 & 31.4 & \textbf{28.86} & 68.42 \\
    Wan2.2-TI2V-5B~\citep{wan2025} & 41.5 & 25.2 & 24.53 & 69.20 \\ 
    Wan2.2-TI2V-5B$^*$~\citep{wan2025} & 64.7 & 28.6 & 24.53  & 69.20 \\ 
    \rowcolor{gray!15} \multicolumn{5}{l}{\textit{Physics-Focused}} \\ 
    PhyT2V (Round 4)$\dagger$~\citep{xue2025phyt2v} & 61 & \underline{37} & - & -\\ 
    WISA$\dagger$~\citep{wang2025wisa} & 62 & 33 & - & -\\ 
    VideoREPA~\citep{zhang2025videorepa} & 51.9  & 22.4  & 21.02 & \textbf{72.54} \\
    VideoREPA$^*$$\dagger$~\citep{zhang2025videorepa} & \textbf{72.1}  & \textbf{40.1}  & 21.02 & \textbf{72.54} \\
    \rowcolor{blue!15}\videogen{} (Ours) &  ~~~~~~~~~~~47.5\increase{14.5\%} & ~~~~~~~~~~~37.9\increase{50.4\%} & ~~~~~~~~~~~\underline{27.75} \increase{13.1\%} & ~~~~~~~~~\underline{71.74}\increase{2.6\%}\\
    \rowcolor{blue!15}\videogen{}$^*$ (Ours) & ~~~~~~~~~\underline{70.3}\increase{8.7\%}  & ~~~~~~~~~~~\underline{39.4}\increase{37.8\%} & ~~~~~~~~~~\underline{27.75}\increase{13.1\%} & ~~~~~~~~~\underline{71.74}\increase{2.6\%} \\
    \bottomrule
  \end{tabular}
  }
  \label{table:videophy_more}
  \vspace{-0.2cm}
  \end{table}
\begin{table*}[t]
\centering
\caption{\textbf{Text-to-video evaluation on \textbf{VBench-2}. } Best results in \textbf{bold}.  Improvements over base model Wan2.2-TI2V highlighted in \increase{green}. 
}
\vspace{-0.3cm}
\resizebox{0.99\linewidth}{!}{
\begin{tabular}{lcccccc}
\toprule
\textbf{Model}  & \textbf{Total} &\textbf{Creativity}& \textbf{Commonsense} &\textbf{Controllability} &\textbf{Human Fidelity} &\textbf{Physics} \\
\midrule
Wan2.2-TI2V-5B~\citep{wan2025} & 51.57 & \textbf{52.50} & 60.57 & 18.50 & 86.10 & 40.19  \\ 
\rowcolor{blue!10} \videogen{} (Ours) &  ~~~~~~~~~\textbf{51.84}\increase{0.5\%} & 45.51 & ~~~~~~~~~~\textbf{61.43}\increase{1.4\%} & ~~~~~~~~~~\textbf{20.23}\increase{9.4\%} & ~~~~~~~~~\textbf{88.39}\increase{2.7\%} & ~~~~~~~~~~\textbf{43.61}\increase{6.0\%} \\ \midrule
\textbf{Model}& \textbf{Human Anatomy} &
\textbf{Human Clothes} &
\textbf{Human Identity} &
\textbf{Composition} &
\textbf{Diversity} &
\textbf{Mechanics} \\ \midrule
Wan2.2-TI2V-5B~\citep{wan2025} & 87.32 & 92.31 & \textbf{78.70} & 40.35 & \textbf{64.67 }& 59.13  \\ 
\rowcolor{blue!10} \videogen{} (Ours) & ~~~~~~~~~\textbf{90.19}\increase{3.3\%} & ~~~~~~~~~\textbf{96.85}\increase{4.9\%} & 78.12 & ~~~~~~~~~~~~~\textbf{45.07 }\increase{11.7\%}  & 45.95 & ~~~~~~~~~~\textbf{60.48}\increase{2.3\%}\\ \midrule
\textbf{Model}& \textbf{Material} &
\textbf{Thermotics} &
\textbf{Multi-view} &
\textbf{Dynamic Spatial Rel.} &
\textbf{Dynamic Attribute} &
\textbf{Motion Order} \\ \midrule
Wan2.2-TI2V-5B~\citep{wan2025} &  36.49 & 54.11 & 11.05 & 24.64 & \textbf{9.52} & 10.77  \\ 
\rowcolor{blue!10} \videogen{} (Ours) & ~~~~~~~~~\textbf{37.33}\increase{2.3\%} 
& ~~~~~~~~~\textbf{54.61}\increase{0.9\%} 
& ~~~~~~~~~~~\textbf{22.01}\increase{99.2\%} 
& ~~~~~~~~~~~\textbf{32.37}\increase{31.4\%} 
& 6.23 
& ~~~~~~~~~~~\textbf{12.46}\increase{15.7\%}\\ \midrule
\textbf{Model}& \textbf{Human Interact.} &
\textbf{Complex Landscape} &
\textbf{Complex Plot} &
\textbf{Camera Motion} &
\textbf{Motion Rationality} &
\textbf{Instance Preservation}\\ \midrule
Wan2.2-TI2V-5B~\citep{wan2025} &   37.33 & \textbf{18.89 }& 9.52 & \textbf{18.83} & 27.59 & \textbf{93.57}\\ 
\rowcolor{blue!10} \videogen{} (Ours) & ~~~~~~~~~~~\textbf{47.00}\increase{25.9\%} 
& 18.22 
& ~~~~~~~~~~\textbf{10.23}\increase{7.5\%} 
& 15.12 
& ~~~~~~~~~\textbf{29.89}\increase{8.3\%} 
& 92.98\\

\bottomrule
\end{tabular}}
\label{tab:vbench2_aug}
\vspace{-0.2cm}
\end{table*}

\noindent \textbf{Quantitative Results.}
Table~\ref{table:videophy_more} presents extended results on VideoPhy~\citep{bansalvideophy} and VideoPhy2~\citep{bansal2025videophy}, including both the evaluation on original prompt and detailed prompts (denoted by $^*$) following VideoREPA~\citep{zhang2025videorepa}.
Across both settings, \videogen{} yields substantial performance gains over the base Wan2.2-TI2V-5B model, demonstrating improved physical commonsense and semantic fidelity. The improvements are especially pronounced under the original-prompt setting, where no dense textual description is provided, indicating that \videogen{} has learned strong intrinsic physics-awareness without relying on enriched prompts.
Despite the fact that VideoREPA~\citep{zhang2025videorepa} is built upon CogVideoX-5B, a considerably stronger backbone than Wan2.2, \videogen{} still delivers large improvements over its base model and achieves competitive performance, underscoring the effectiveness of our approach.

Table \ref{tab:vbench2_aug} reports fine-grained performance across all 18 VBench-2 metrics. \videogen{} outperforms the base Wan2.2-TI2V-5B on the majority of these dimensions, demonstrating that joint physics-aware modeling not only boosts physics-related metrics but also helps improve overall perceptual realism, semantic consistency, and temporal coherence.

\begin{figure*}[!t]
    \centering
    \includegraphics[width=0.99\linewidth]{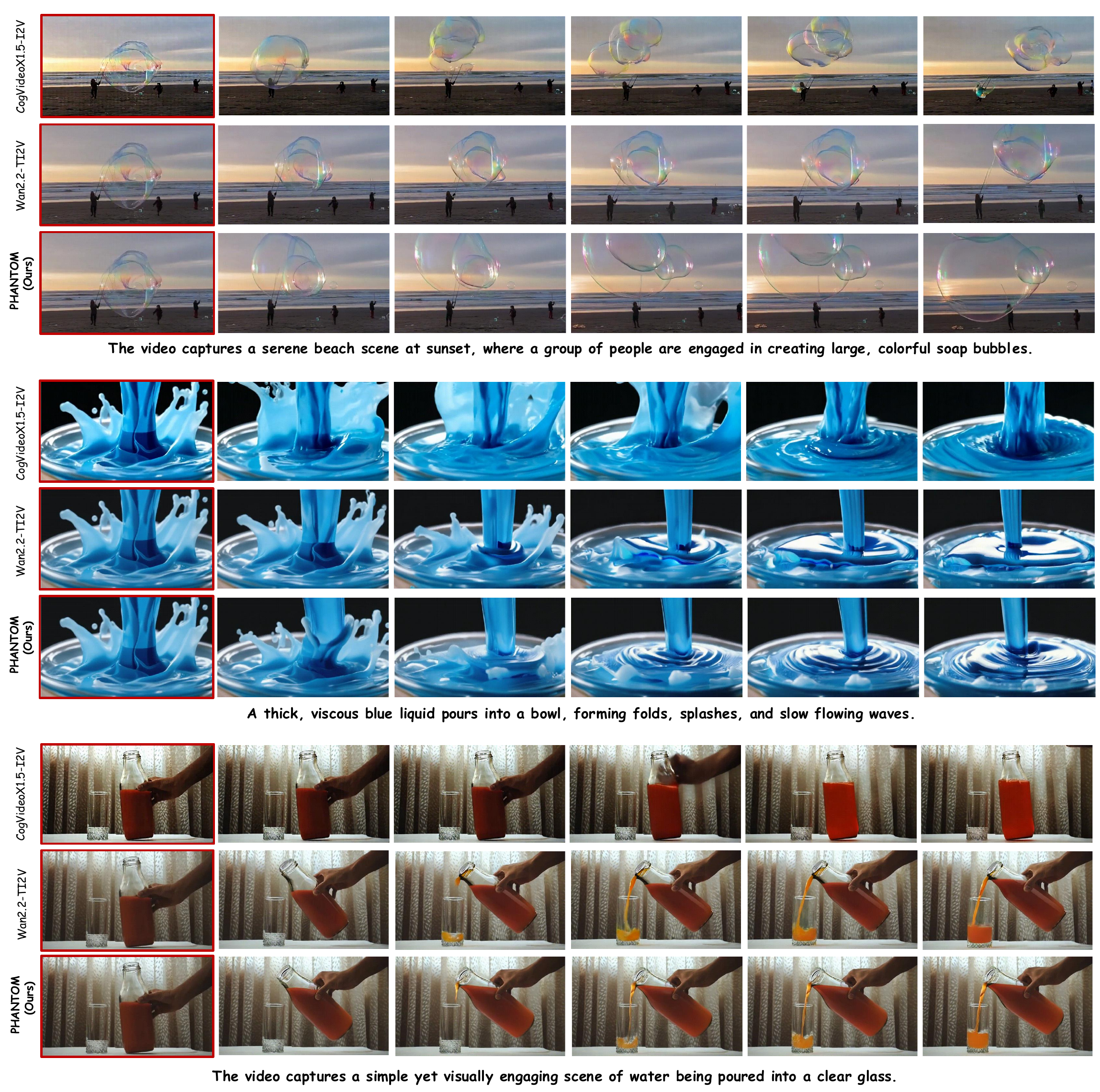}
    \vspace{-0.3cm}
    \caption{\textbf{Qualitative Comparison on Text-/Image-to-Video Generation.} The conditional frame is marked in red box.}
    \label{fig:ti2v}
\end{figure*}
\begin{figure*}[!t]
    \centering
    \includegraphics[width=\linewidth]{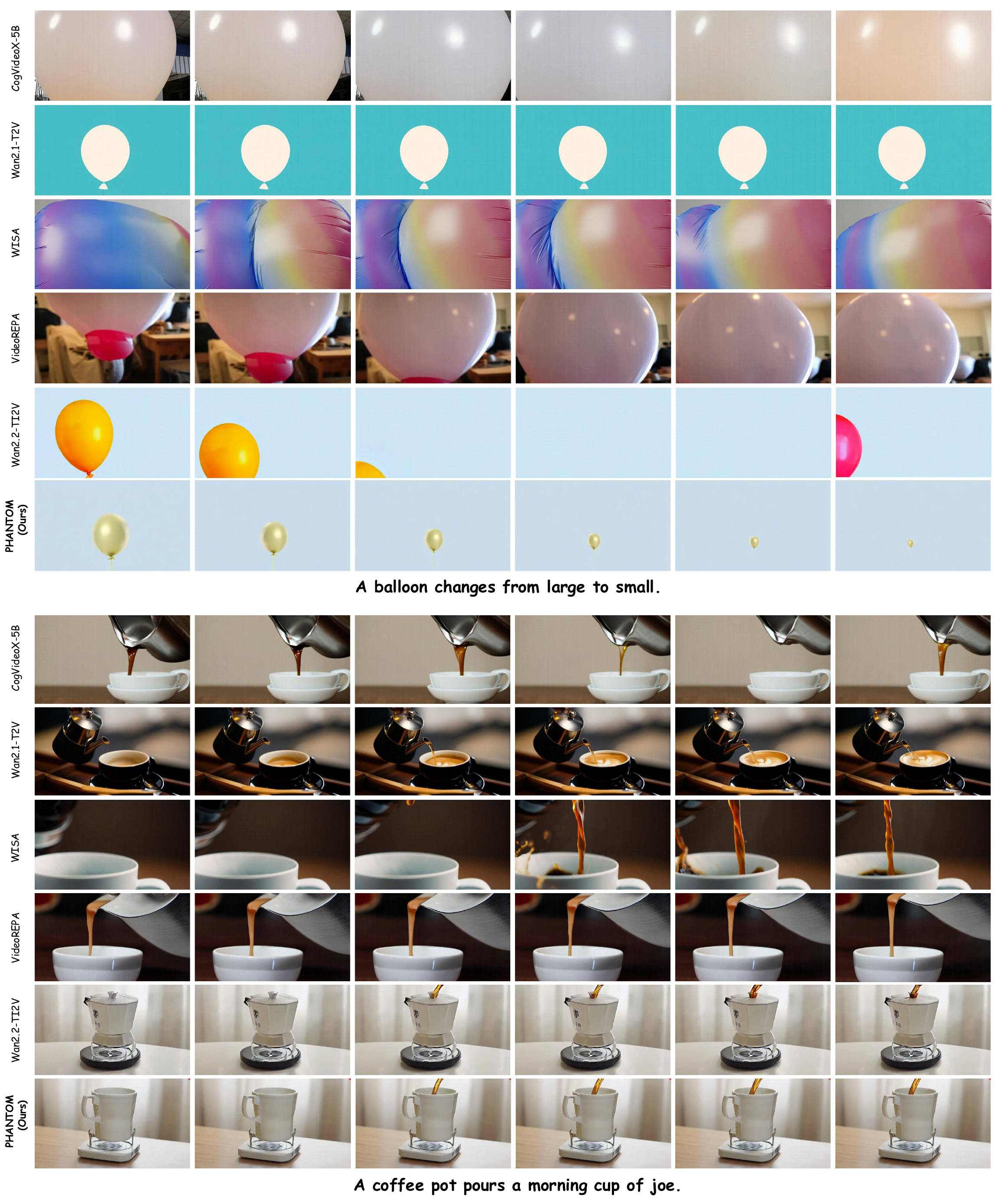}
    \vspace{-0.3cm}
    \caption{\textbf{Qualitative Comparison on Text-to-Video Generation.}}
    \label{fig:t2v}
\end{figure*}

\noindent \textbf{Ablation Studies.}
In addition, we replace the VJEPA2 encoder with VideoMAEv2, an alternative video encoder, while keeping the same training setup on Wan2.2-TI2V. \Cref{table:videophy_reb} shows \videogen{} w/ VJEPA2 achieves better performance across all metrics, supporting the choice of VJEPA2 for physics-aware latent representation.

\noindent \textbf{Qualitative Results.}
We provide additional qualitative comparisons against both state-of-the-art T2V models and recent physics-focused approaches, as shown in Figure \ref{fig:ti2v} and \ref{fig:t2v}.
Since most physics-focused baselines operate solely in the text-to-video setting, Figure~\ref{fig:t2v} compares \videogen{} only with general-purpose T2V models.

\section{Physics-based Video Control}

To further evaluate the ability of \videogen{} to model and respond to explicit physical control signals, we apply our framework to the Force-Prompting dataset \footnote{\url{https://force-prompting.github.io}}.
Force-Prompting provides paired video sequences and temporally aligned force annotations describing external physical interactions applied to static images.
Specifically, we focus on the local point force setting, in which a localized force is applied to an object at a specific image coordinate.

We convert each point-force annotation into a \emph{force tensor} that encodes both the spatial distribution and temporal evolution of the applied forces.
Each tensor is then rendered as a short video sequence at a resolution of $256 \times 256$, providing a consistent spatiotemporal representation of the external force.
These force videos are processed by the V-JEPA2 encoder in the same way as ordinary video inputs, producing physics-aware embeddings that are fed through the physics branch.\looseness-1

Since the original video captions in the Force-Prompting dataset do not contain force-related information, we additionally construct a textual \emph{force prompt} that describes the applied force in natural language. This prompt encodes all relevant physical parameters and is fed into the physics branch during training and inference:
\vspace{-0.2cm}
\begin{verbatim}
Simulate the scene under an external point 
force applied at (x, y) = ({coordx}, {coordy}), 
with magnitude = {force} and direction = {angle} 
degrees, and generate the resulting video dynamics.
\end{verbatim}
\vspace{-0.2cm}

In this application, the two branches of \videogen{} receive different inputs and textual conditions. 
The \emph{video branch} models the visual evolution of the scene and is conditioned on the original video caption.
In contrast, the \emph{physics branch} processes the force-tensor video and is guided by the constructed force prompt. During inference, \videogen{} is conditioned on a single static image along with the first frame of the force-tensor sequence, and it synthesizes the resulting physically driven dynamics.

\begin{table}[t!]
  \centering
  \caption{\textbf{Alternative Video Encoders.} 
  }
  \vspace{-0.3cm}
  \resizebox{0.92\columnwidth}{!}{%
  \begin{tabular}{l  c c c c}
    \toprule
    \textbf{Visual Encoder} & \multicolumn{2}{c}{\textbf{VideoPhy}} & \multicolumn{2}{c}{\textbf{VideoPhy-2}} \\ \cmidrule(lr){2-3} \cmidrule(lr){4-5} 
    & \textbf{SA}$\uparrow$ & \textbf{PC}$\uparrow$  &  \textbf{SA}$\uparrow$ & \textbf{PC}$\uparrow$  \\ \midrule
    Wan2.2-TI2V-5B~ & 41.5 & 25.2 & 24.53 & 69.20 \\
   \rowcolor{blue!10} \videogen{} w/ VJEPA-2 &  47.5 & 37.9 & 27.75 & 71.74  \\
    \rowcolor{blue!10} \videogen{} w/ VideoMAEv2 & 45.8 & 37.6 & 26.90 & 70.56 \\
    \bottomrule
  \end{tabular}
  }
  \label{table:videophy_reb}
  \vspace{-0.3cm}
  \end{table}
  
\begin{figure*}[!t]
    \centering
    \includegraphics[width=\linewidth]{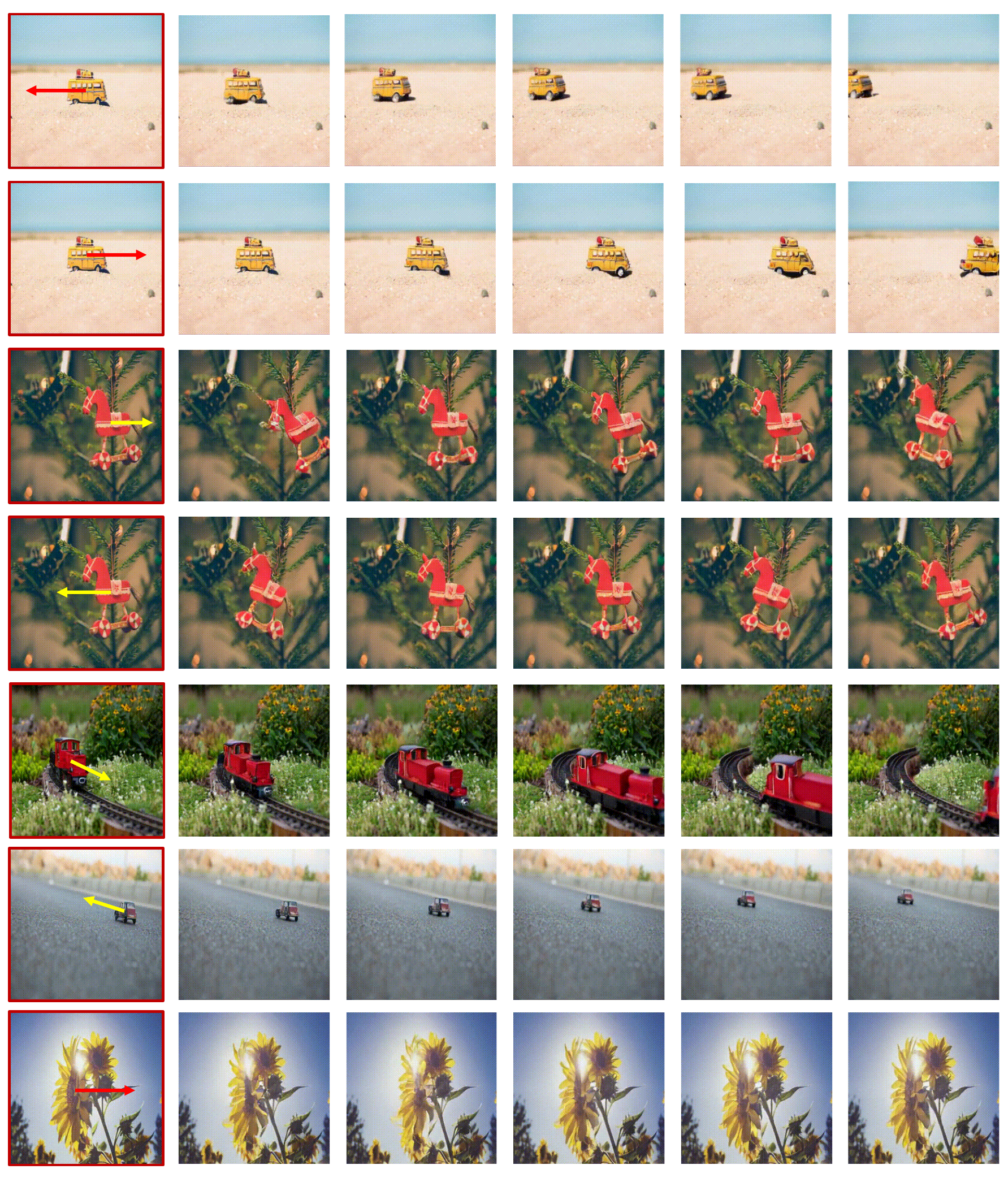}
    \vspace{-0.5cm}
    \caption{\textbf{Examples of Force-conditioned Video Generation using \videogen{}.} The conditional frame is marked in red box.}
    \label{fig:force}
\end{figure*}

We follow the same experimental hyperparameters as in our main setup and fine-tune from the \videogen{} for 1.1K steps.
Figure \ref{fig:force} shows that \videogen{} can synthesize dynamic and physically plausible motion that evolves consistently with the applied force, demonstrating its ability to generalize force-based control signals.

\end{document}